\documentclass[10pt,journal,onecolumn]{IEEEtran}
\usepackage{setspace}
\ifCLASSINFOpdf
\else
\fi
\usepackage{amssymb}

\usepackage{amsfonts,amsmath,amssymb,latexsym,float,epsfig,subfigure,epstopdf,graphicx,array,algorithm,algpseudocode}
\usepackage{hyperref}

\begin{document}
	%
	\title{Diffusion-KLMS Algorithm and its Performance Analysis for Non-Linear Distributed Networks\footnote{This paper is a preprint of a paper submitted to IET Signal Processing (special issue on 5G wireless networks) and is subject to Institution of Engineering and Technology Copyright. If accepted, the copy of record will be available at IET Digital Library}}
	%
	%
	%
	
	\author{Rangeet~Mitra,~\IEEEmembership{}
		and~Vimal~Bhatia,~\IEEEmembership{Senior Member,~IEEE}
	}
	
	%
	%

	\markboth{Special Issue on 5G wireless networks}%
	{Shell \MakeLowercase{\textit{et al.}}: Bare Demo of IEEEtran.cls for Journals}
	%



	\maketitle
	
	\begin{abstract}
		\\In a distributed network environment, the diffusion-least mean squares (LMS) 
		algorithm gives faster convergence than the original LMS algorithm. It has also been observed that, the diffusion-LMS generally outperforms other distributed LMS algorithms like spatial LMS and incremental LMS. However, both the original LMS and diffusion-LMS are not applicable in non-linear environments where data may not be linearly separable. 
		A variant of LMS called kernel-LMS (KLMS) has been proposed in the literature 
		for such non-linearities. In this paper, we propose kernelised version of diffusion-LMS for non-linear distributed environments. Simulations show that the proposed approach has superior convergence as compared to algorithms of the same genre. We also introduce a technique to predict the transient and steady-state behaviour of the proposed algorithm. The techniques proposed in this work (or algorithms of same genre) can be easily extended to distributed parameter estimation applications like cooperative spectrum sensing and massive multiple input multiple output (MIMO) receiver design which are potential components for 5G communication systems.
	\end{abstract}
	

	%
	\IEEEpeerreviewmaketitle

	\section{Introduction}
	Nowadays, there is a thrust toward development of a new standard for communications called 5G, which involves some novel approaches like massive multiple input multiple output (MIMO), cooperative spectral sensing, visible light communication (VLC) etc. \cite{boccardi2014five}. Massive MIMO uses a large number of antenna array elements (which consist of antennae at the receiver and those at the network nodes) which greatly increases the capacity of the communication system. Spectral sensing is a technique to estimate vacant spectral subbands adaptively. Such vacant subbands may be used to accommodate incoming transmission which saves bandwidth as we are saved from  allocating a new frequency band for the incoming signal. The distributed diffusion based adaptive filtering algorithms have potential applications in cooperative spectral sensing and distributed MIMO detection
	\cite{li2012adaptive,wang2014cellular,cattivelli2011distributed}. Hence distributed adaptive filtering/optimization over distributed networks is an important and emerging research area which can be applied to 5G standard components.
	
	Distributed signal processing deals with drawing inferences from data coming from various nodes in a given graph. Robust distributed algorithms are required to draw inferences from the intelligently fused data from all the nodes. The task of training an artificial computer to automatically draw inferences and take decisions is assigned to the statistical learning techniques. Statistical learning algorithms may be categorised into four distinct classes: a) Supervised learning, b) Unsupervised learning, c) Semi-supervised learning and d) Reinforcement learning \cite{alpaydin2004introduction}. In supervised learning, the data labels are assumed to be known during training. In unsupervised learning, the data labels are not known while training. In semi-supervised learning, only a subset of the labels are known. In reinforcement learning, the algorithm is trained in such a way so as to maximise a utility function. The scope of this paper is limited to distributed supervised learning.
	
	One of the well known supervised learning rules is the Widrow-Hoff learning rule or the least mean squares (LMS) algorithm. It belongs to the class of stochastic gradient algorithms. It replaces the expectation operator in the Weiner-Hopf equation \cite{hayes2009statistical} by the instantaneous gradient of the quadratic cost function. In the recent literature \cite{cattivelli2010diffusion,lopes2007incremental}, there has been a major thrust towards generalising the LMS algorithm in distributed environments. A variant of the widely known LMS algorithm or the Widrow-Hoff learning rule, called the diffusion-LMS, has been used in distributed optimisation in \cite{cattivelli2010diffusion} with wide number of application areas. This algorithm uses stochastic matrices to fuse the data intelligently coming from different sources (for example, nodes of the network) and has the best performance among all distributed counterparts of LMS algorithm \cite{zhao2012performance,sayed2014adaptive}. Similarly other extensions of adaptive filtering algorithms like recursive least squares (RLS) called diffusion-RLS have also been proposed \cite{cattivelli2008diffusion}.

	Classical adaptive filtering algorithms like LMS and diffusion-LMS (for networks) work well for affinely separable data. However, in scenarios when the data is not guaranteed to be affinely separable \cite{alpaydin2004introduction}, which occurs frequently in non-linear scenario, the kernel least mean squares (KLMS) algorithm has been found in the literature to perform better as demonstrated in \cite{liu2008kernel} and has found wide applicability as in \cite{haghighat2015variable,fan2010proportional,raja2014adaptive}. The basic principle of KLMS is the kernel trick \cite{alpaydin2004introduction}, which maps the input data into a linearly separable high dimensional reproducing kernel Hilbert space (RKHS) \cite{liu2008kernel}. Similar extensions to linear algorithms like affine projection algorithm to kernel spaces exist as in \cite{slavakis2008sliding}.
	Kernel based distributed learning algorithms have been proposed in the literature \cite{predd2007distributed,honeine2008distributed,honeine2009functional}. However, they neither address the kernel LMS regression problem in  the diffusion framework nor is their performance analysed in terms of popular performance metrics.
	
	In this paper, we propose an extension of KLMS for distributed networks. In other words, we seek to apply the kernel trick to the diffusion-LMS adaptations given in \cite{sayed2014adaptive}. We also seek to provide theoretical expressions that govern the proposed algorithm's transient and steady state behaviour as has been done in \cite{sayed2014adaptive,khalili2012steady} by classical adaptive filtering theory based approaches as given in \cite{diniz1997adaptive}.
	
	This paper is organised as follows: to facilitate understanding of background material and concepts forming theoretical basis of the proposed algorithm, the diffusion-LMS algorithm and KLMS algorithm  are reviewed in Section-II and Section-III respectively. The diffusion KLMS algorithm is proposed in section-IV. To gain insights into the performance of the algorithm, transient performance, steady-state performance and condition for convergence are mathematically analysed in Section-V. The simulation results and comparison with other algorithms is provided in Section-VI, and Section-VII concludes the paper.
	\section{Review of distributed Diffusion LMS}
	In this section, we review the distributed diffusion-LMS given as given in \cite{sayed2014adaptive}. In the distributed diffusion-LMS algorithm, there are a set of nodes in a graph $\mathcal{G}$. The neighbourhood of a node in a graph is given by a set of nodes $\mathcal{G^{'}}$ such that there exists an edge between that node and the nodes in the set $\mathcal{G^{'}}$. Please note that for each node, $\mathcal{G^{'}}$ also includes the node itself. Let stochastic matrices be given by the entries ${A}=[a_{ij}]$ and ${C}=[c_{ij}]$ represent a probabilistic weight from node $i$ to node $j$. This matrix is generally determined by stochastic sampling techniques as given in \cite{sayed2014adaptive}.
	
	For a distributed adaptive graph indexed by time variable $n$, the adaptive filter attempts to estimate the local cost $J_{q}(n)$ function at time instant $n$ at a given node:
	\begin{equation}
	J_{q}(n)=\sum_{l\varepsilon \mathcal{G^{'}}}c_{lq}J_{l}(n)
	\end{equation}
	where $l$ runs over all members of the neighbourhood of the $q^{th}$ node of the network and forms the $q^{th}$ local cost function $J_{q}$.
	For this, the distributed Weiner solution based local estimate at node $q$ will be, ${w}_{q}^{o}$, and is given as:
	\begin{equation}
	{w}^{o}_{q}(n) = (\sum_{l\varepsilon \mathcal{G^{'}}}c_{lq}R_{x_{ln}})^{-1}(\sum_{l\varepsilon \mathcal{G^{'}}}c_{lq}r_{dx_{ln}})
	\end{equation}
	where, $R_{x_{ln}}$ is the autocorrelation matrix for the $l^{th}$ node in the neighbourhood of the node $q$ of the graph. $r_{dx_{ln}}$ is the cross correlation between the desired output $d$ and ${x}_{l}$ is the data from $l^{th}$ member of the neighbourhood of node $q$ at time $n$.
	
	The weight vector ${w}_{q}$, for the $q^{th}$ node, is iteratively adapted by diffusion-LMS as follows,
	\begin{equation}\label{step1}
	{p}_{q}(n+1) = {p}_{q}(n) + \mu\sum_{l\varepsilon\mathcal{G^{'}}}c_{lq}(d_{l}(n)-{w}_{l}(n)^{T}{x}_{l}){x}_{l}
	\end{equation}
	\begin{equation}\label{step2}
	{w}_{q}(n+1) = \sum_{l\varepsilon \mathcal{G^{'}}} a_{lq}{p}_{l}(n+1)
	\end{equation}
	where, $\mu$ is the step-size, $d_{l}(n)$ is the desired response at the $l^{th}$ node at the $n^{th}$ time instant and ${p}_{q}(n)$ is the vector of intermediate value of the adaptive filter at $q^{th}$ node at $n^{th}$ time instant before it can be combined probabilistically over its neighbourhood to get the final updated estimate.

	The steps in eq. (\ref{step1}) and (\ref{step2})  can be carried out in either order. In both situations, it will belong to the same genre of algorithms. If the eq. (\ref{step1}) is carried out first it is called Adapt and Then Combine (ATC) diffusion. If the eq. (\ref{step2}) is carried out first it is called Combine and Then Adapt (CTA) diffusion \cite{sayed2014adaptive}.
	
	Please note that an important factor in convergence of the adaptive filters is the spectral radius of the covariance matrix. This spectral radius is a norm in itself. Applying Jensen's inequality to the spectral radius as in \cite{sayed2014adaptive}, $\rho_{max}$ of the weighted covariance matrix,
	\begin{equation}
	\rho_{max}(\sum_{l\varepsilon \mathcal{G^{'}}}c_{lq}R_{lq}) \leq \sum_{l\varepsilon \mathcal{G^{'}}}c_{lq}\rho_{max}(R_{lq}) \leq \max_{1\leq l\leq N} \rho(R_{lq})
	\end{equation}
	where, $N=|\mathcal{G}^{'}|$ and $R_{lq}$ is the autocorrelation matrix of the $l^{th}$ neighbour of the $q^{th}$ node.
	Hence, due to lower eigen-value spread, it converges faster. More rigorous convergence results are found in \cite{sayed2014adaptive}.
	
	\section{Review of KLMS}
	The linear-LMS as described in \cite{alpaydin2004introduction,hayes2009statistical} minimises the following cost function at $n^{th}$ instant:
	\begin{equation}
	J_{LMS}(n) = \mathbb{E}[(d(n)-{w}(n)^{T}{x}_{n})^2]
	\end{equation}
	where ${x}_{n}$ is the observation vector for the $n^{th}$ time instant and $\mathbb{E}\lbrack\cdot\rbrack$ is the expectation operator.
	Dropping the expectation operator and taking gradient with respect to ${w}$, we arrive at the following stochastic gradient update rule \cite{alpaydin2004introduction},
	\begin{equation}
	{w}(n+1) = {w}(n) + \mu e_{LMS}(n) {x}_{n}
	\end{equation}
	where,
	$e_{LMS}(n)=(d(n)-{w}(n)^{T}{x}_{n})$
	
	When the data is not linearly separable the above adaptation does not converge to optimum value. Hence, in such scenarios, we invoke the kernel trick and map the vectors to RKHS as in \cite{alpaydin2004introduction}  by a feature map $\phi:\mathbb{R}^{m}\to\mathcal{H}$.
	
	In RKHS, the adaptation can be written as follows:
	
	\begin{equation}
	\Omega(n)=\Omega(n-1)+\mu e_{KLMS}(n-1)\phi(x_{n-1})
	\end{equation}
	where $\Omega$ is the implicit parameter to be estimated in RKHS.
	This can be written as a running summation as follows:
	
	\begin{equation}
	\Omega(n)=\mu \sum_{i=0}^{n-1}e_{KLMS}(i)\phi(x_{i})
	\end{equation}
	
	Taking inner product with the latest observation and assumption of zero initial conditions would give the following recursion as in \cite{liu2008kernel}:
	\begin{equation}
	y(n+1) = \mu \sum_{i=0}^{k-1}e_{KLMS}(i)<\phi({x}_{i}){,}\phi({x}_{n})>_{\mathcal{H}}
	\end{equation}
	
	where,
	
	\begin{equation}
	e_{KLMS}(n) = (d(n) - y(n))
	\end{equation}
	is the error at $n^{th}$ instant and $<\cdot{,}\cdot>_{\mathcal{H}}$ denotes a real kernel inner product \cite{liu2008kernel} on RKHS $\mathcal{H}$. Several possibilities of kernel inner products exist; some of them being polynomial and Gaussian kernels \cite{alpaydin2004introduction}.
	This algorithm has a nice self-regularising property, and has been studied in details in \cite{liu2008kernel}.

	\section{Proposed Diffusion-KLMS}
	Based on the KLMS algorithm, reviewed in the previous section, we propose its distributed variant in this section based on the diffusion approach. We now define matrices and symbols that will be used in this paper.
	In this proposal, we have the matrix $Y=[y(l,n)]$ to denote output corresponding to the $l^{th}$ neighbour at $n^{th}$ time instant.
	$E=[e(l,n)]$ is the error matrix corresponding to the $l^{th}$ neighbour at $n^{th}$ time instant.
	$X=[\{{x}_{l}(n)\}]$ is a matrix of measurement vectors  from neighbours of node $q$ at time instant $n$ stacked together.
	In the following few lines, we will denote the collection of the data from various nodes at the $n^{th}$ time instant as $X(n)$. $X(n)$ contains the data pertaining to all $l$ neighbours stacked in row vector form. In case, there is no vector from a node in the neighbourhood it is replaced by the zero vector in $X(n)$ and will have a corresponding 0 entry in ${C}$.
	
	
	

	The gradient from eq. (\ref{step1}) is redefined as:
	\begin{equation}
	\nabla_{{p}_{q}}J_{q}(n) =  e(l,n)^{'}\phi(CX(n))
	\end{equation}
	where $\phi(.)$ is a feature map from $\mathbb{R}^{d}\to\mathcal{H}$, where $d$ is the dimensionality of the data and $\mathcal{H}$ is an RKHS.
	Applying the kernel trick results in,
	\begin{equation}\label{algo2}
	y(l,n+1) = \mu\sum_{i=0}^{n-1} e(l,n)^{'}<CX(i),X(n)>_{\mathcal{H}}
	\end{equation}
	\begin{equation}\label{algo1}
	e(q,n+1)^{'} = \sum_{l\varepsilon \mathcal{G^{'}}} a(q,l) d_{l}(n)-  \sum_{l\varepsilon \mathcal{G^{'}}} a(q,l) y(l,n)
	\end{equation}
	where $A$ is a stochastic matrix corresponding to the probabilistic weights $\{a(q,l)\}$.
	The error at $n^{th}$ time instant at the $q^{th}$ node would be the (transformed) mean (by A) of $e$ over all possible $l$.
	
	The proposed algorithm is given below, as iterating following three steps, till convergence:
	\begin{enumerate}
		\item Estimate the outputs of node $l$ using estimates of error $e_{l}^{'}(n)$.
		\item Form an estimate of errors at time instant $n$ at each node $l$. Let this be given by the vector ${e}(n)$ whose $l^{th}$ element is $e(l,n)$. Then the error term for the $l^{th}$ node for the $n^{th}$ time instant can be written as $e(l,n) = d(n) - y(l,n)$
		\item The error at each node is modified by the transformation $A$ by the equation ${e}^{'}(n+1)=Ae(n)$,
		where $e(n)$ and $e^{'}(n)$ are vectors of error terms corresponding to all the nodes (for all nodes indexed by $l$) stacked together.
	\end{enumerate}
	
	\section{Transient and Steady State Performance}
	
	In this section, we provide the steady state analysis of the proposed algorithm based on the classical approach outlined in \cite{diniz1997adaptive} (analysis based on eigenvalues of autocorrelation matrices). We note that the proposed recursion for the $q^{th}$ node can be expressed in RKHS as follows:
	
	\begin{gather}\label{RKHS1}
	\Omega_{q}(n)=\Omega_{q}(n-1)-\sum_{\forall l}e_{q}(n)c_{lq}\phi(x_{l})\\ \nonumber
	y_{q}(n)=<\Omega_{q}(n),\phi(x_{obs})>_{\mathcal{H}} \\ \nonumber
	e_{q}(n)=d_{q}(n)-y_{q}(n) \\ \nonumber
	e_{q}(n+1)=\sum_{\forall l}a_{lq}e_{l}(n)
	\end{gather}
	$\Omega_{q}$ is an implicit parameter which is learned in RKHS and $x_{obs}$ is an input observation. Let the optimal value of the parameter be $\Omega^{o}$
	and the deviation of the implicit parameter from the optimal value in $q^{th}$ node at $n^{th}$ instant be denoted as $\tilde{\Omega}_{q}(n)$. Subtracting $\Omega^{o}$ from both sides of first equation of (\ref{RKHS1}), we get:
	\begin{gather}
	\tilde{\Omega}_{q}(n)=\tilde{\Omega}_{q}(n-1)-\sum_{\forall l}e_{q}(n)c_{lq}\phi(x_{l})
	\end{gather}
	Taking inner product on both sides of the above equation with $\phi(x_{l})$,
	\begin{gather}\label{RKHS2}
	\tilde{y}_{q}(n)=\tilde{y}_{q}(n-1)-\mu\sum_{\forall l}c_{lq}(\tilde{y}_{q}(n-1)+n_{q})<\phi(x_{l}){,}\phi(x_{obs})>_{\mathcal{H}}\\ \nonumber
	=(1-\mu\sum_{\forall l}c_{lq}<\phi(x_{l}){,}\phi({x}_{obs})>_{\mathcal{H}})\tilde{y}_{q}(n-1) - \mu \sum_{l}c_{lq}n_{q}<\phi(x_{l}){,}\phi({x}_{obs})>_{\mathcal{H}}
	\end{gather}

    Please note that $\tilde{y}_{q}(n)$ is calculated after combination by the $A$ matrix in the last step of eq. (\ref{RKHS1}).
	Define a matrix $A_{1} = A \otimes I_{D}$ and $C_{1}= C \otimes I_{D}$, where $A$ and $C$ are combining matrices and
	$I_{D}$ is a $D\times D$ identity matrix; where $D$ is the cardinality of the network. Further we define two vectors namely, $\Phi(x) = [\phi(x_{1}),\phi(x_{2}),...,\phi(x_{d})]^{T}$
	and $\Phi(x_{obs})=[\phi(x_{obs}),\phi(x_{obs}),...,\phi(x_{obs})]^{T}$.
	Using above defined variables, we rewrite (\ref{RKHS2}) as,
	\begin{gather}
	\tilde{y}_{q}(n)=(1-\mu<C_{1}\Phi(x),\Phi(x_{obs})>_{\mathcal{H}})\tilde{y}_{q}(n-1)-\mu <C_{1}\Phi(x),\Phi(x_{obs})>_{\mathcal{H}}n_{q}
	\end{gather}
	Squaring both sides, taking expectation, and considering only till the first power of $\mu$, we get:
	\begin{gather}\label{transient}
	\mathbb{E}[|\tilde{y}_{q}(n)|^2]=[1-2\mu<C_{1}\Phi(x),\Phi(x_{obs})>_{\mathcal{H}}]\mathbb{E}[|\tilde{y}_{q}(n-1)|^2]+\mu^2\sigma_{n}^{2} \mathbb{E}(|<C_{1}\Phi(x),\Phi(x_{obs})>_{\mathcal{H}}|^2)
	\end{gather}
	Based on (\ref{transient}) we derive the transient behaviour, steady state behaviour and condition for convergence of the proposed algorithm.
	
	\subsection{Transient behaviour}
	To estimate the speed of convergence of the proposed approach it is essential to gain insight into the dynamical equation that governs the evolution of the learning curve vs number of iterations.
	
	The above dynamical equation (\ref{transient}) controls the transient behaviour at small step-sizes. The inner product $<\phi(x){,}\phi(y)>_{\mathcal{H}}$ depends on choice of kernel. As we use a real Gaussian kernel as done in \cite{liu2008kernel},
	\begin{equation}\label{kernel_exp}
	<\phi(x){,}\phi(y)>_{\mathcal{H}}=\frac{1}{\sqrt{2\pi\sigma^2}}\exp(-\frac{\|x-y\|^2}{2\sigma^2})
	\end{equation}
	where $x,y\in \mathbb{R}^{m}$ and $\phi:x\to\phi(x)$ is a feature map from the vector space of real numbers to RKHS.
	Using the definition of $<\cdot{,}\cdot>_{\mathcal{H}}$ given in (\ref{kernel_exp}) in (\ref{transient}) we get the transient behaviour of the proposed approach. We see that for a given $\mu$ and noise variance $\sigma_{n}^2$ the transient behaviour of the proposed approach can be easily modeled using (\ref{transient}).
	\subsection{Steady state behaviour}
	It is also essential to see how the MSE floor to which the proposed algorithm has converged varies with step-size.
	From (\ref{transient}), assuming convergence ($\mathbb{E}[|\tilde{y}_{q}(n)|^2] \approx \mathbb{E}[|\tilde{y}_{q}(n-1)|^2]$), we arrive at the following expression for misadjustment,
	\begin{equation}
	\mathbb{E}[|\tilde{y}_{q}(n)|^2]= \frac{\mu \sigma_{n}^2}{2}\mathbb{E}(|<C_{1}\Phi(x),\Phi(x_{obs})>_{\mathcal{H}}|)
	\end{equation}
	Thus we can see that the above equation (derived for non-linear systems) is similar to equation derived in \cite{diniz1997adaptive} for the Widrow-Hopf learning rule for single node for linear parameter estimation.
	\subsection{Step-size range for convergence}
	For any adaptive algorithm it is very important to set up the step-size, $\mu$, in the range in which the algorithm converges. If $\mu$ is too less, we may observe slow convergence. Too high a $\mu$ may result in mis-convergence.
	
	The proposed algorithm converges iff the following condition holds,
	\begin{gather}
	[1-2\mu<C_{1}\Phi(x),\Phi(x_{obs})>_{\mathcal{H}}+\mu^2|<C_{1}\Phi(x),\Phi(x_{obs})>_{\mathcal{H}}|^2]<1\\ \nonumber
	\implies
	0<\mu<\frac{2}{<C_{1}\Phi(x),\Phi(x_{obs})>_{\mathcal{H}}}
	\end{gather}
	Hence, if $\mu$ is in the above range then the proposed algorithm converges. This bound (derived for non-linear systems) is similar to the general case of the bound of the convergence of step-size for Widrow-Hopf learning rule for a single node linear scenario.

	\section{Results}
	
\begin{figure}
		\centering
		\includegraphics[width=10cm,height=9cm]{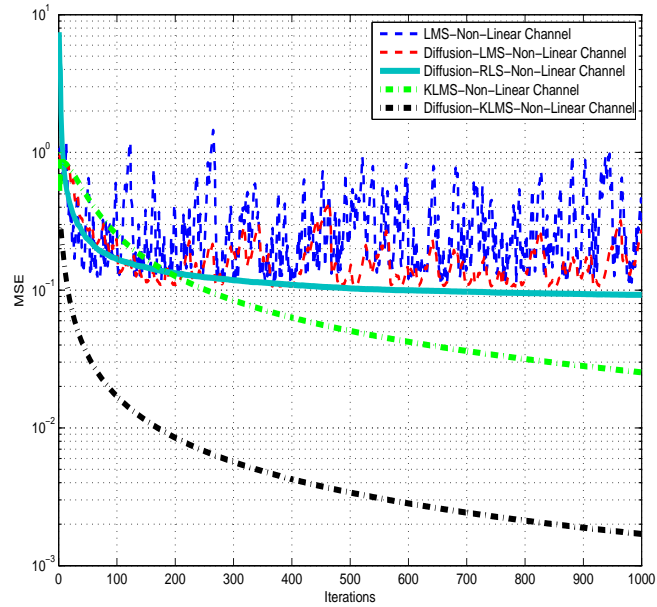}\\
		\caption{Convergence plot for LMS, Diffusion-LMS, Diffusion-RLS, KLMS and Diffusion KLMS: ${A}={C} = [0.5\hspace{3mm} 0.5;0.5\hspace{3mm} 0.5]$}\label{fig2}
	\end{figure}

\begin{figure}[h]
		\centering
		\includegraphics[width=10cm,height=9cm]{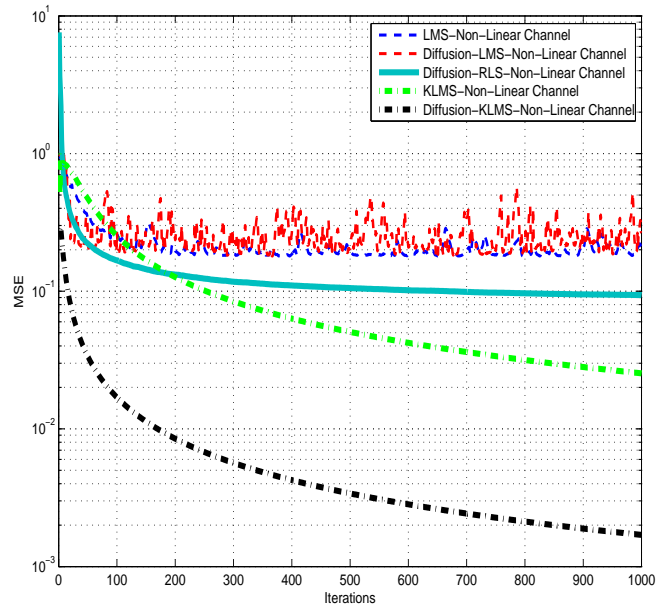}\\
		\caption{Convergence plot for LMS, Diffusion-LMS, Diffusion-RLS, KLMS and Diffusion-KLMS comparison:
			${A}=[0.666\hspace{3mm} 0.333;0.333\hspace{3mm} 0.666]$,${C} = [0.5\hspace{3mm} 0.5;0.5\hspace{3mm} 0.5]$}\label{fig3}
	\end{figure}

	In this section, we present the simulation results based on the analysis presented in previous sections.
	An independently identically distributed (i.i.d) sequence \{$\pm1$\} was generated. Consequently, this  sequence was passed through a non-linearity $f(x)= x-0.9x^2$ as in \cite{liu2008kernel} so as to simulate a non-linear system. Further, additive white Gaussian noise of variance 0.16 was added. In other words, we considered a simple de-noising problem for our simulations. The convergence and error performance of KLMS and diffusion-KLMS are shown in Fig. \ref{fig2} for ${A}={C} = [0.5\hspace{3mm} 0.5;0.5\hspace{3mm} 0.5]$ and in Fig. \ref{fig3} for ${A}=[0.666\hspace{3mm} 0.333;0.333\hspace{3mm} 0.666]$,${C} = [0.5\hspace{3mm} 0.5;0.5\hspace{3mm} 0.5]$. We see that although the LMS and diffusion LMS perform well in linear channels, they fail to converge in non-linear channels.
	We observe superior convergence to lower MSE floors is case of diffusion-KLMS as compared to KLMS, LMS, diffusion LMS and diffusion-RLS. We use $\mu=0.2$ and spread parameter $\sigma=0.1$ for KLMS and the proposed KLMS based approach. For LMS and diffusion-LMS, step-size $\mu =0.02$ is used for simulation.
	We observe performance gain of two decades of the proposed approach with respect to LMS and diffusion-LMS. Also, we find a gain of a decade of performance with respect to single-node KLMS. We observe that the linear RLS exhibits poor performance in a non-linear scenario as the covariance matrix updation fails due to non-linearity.
	
\begin{figure}[!ht]
		\centering
		\includegraphics[width=18cm,height=18cm]{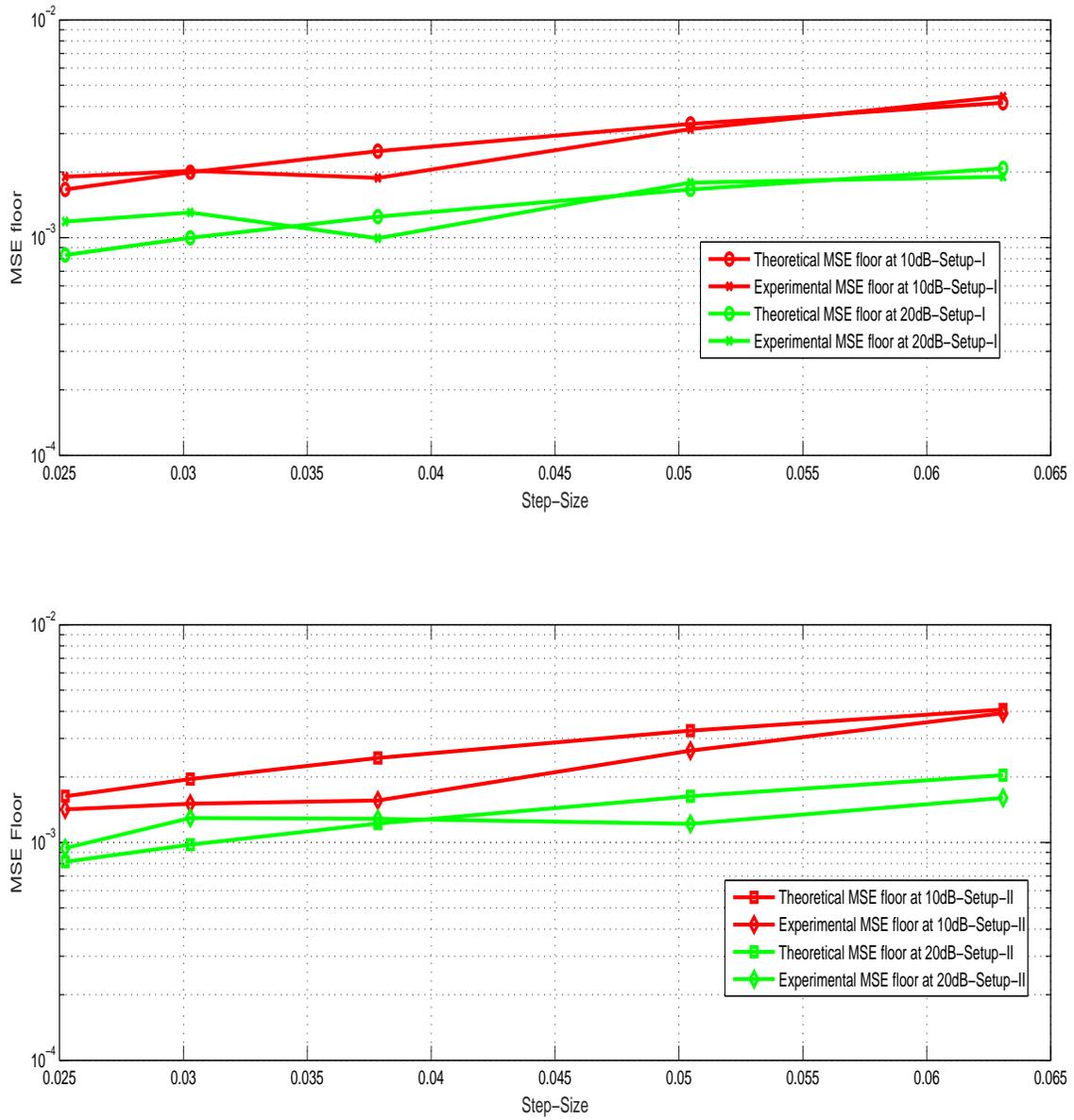}\\
		\caption{ MSE floors comparison for Diffusion-KLMS :Theoretical and Experimental; Setup-I: ${A}=[0.5\hspace{3mm} 0.5;0.5\hspace{3mm} 0.5]$, ${C} = [0.5\hspace{3mm} 0.5;0.5\hspace{3mm} 0.5]$, Setup-II: ${A}=[0.666\hspace{3mm} 0.333;0.333\hspace{3mm} 0.666]$,  ${C} = [0.5\hspace{3mm} 0.5;0.5\hspace{3mm} 0.5]$}\label{fig8}
	\end{figure}

	In Fig. \ref{fig8}, the steady state behaviour of diffusion KLMS as a function of step-size where the theoretical curves, which are obtained from Section-V, are observed to be close to the experimental curves. 
	The computational complexity of training phase of the proposed scheme is $O(D^{2}|\mathcal{G}|)$ and testing computational complexity is $O(D|\mathcal{G}|)$ as the computational complexity of the training and testing phases are given as $O(D^2)$ and $O(D)$ respectively as in \cite{liu2008kernel} where $D$ is the dimensionality of the observations.

\begin{figure}[!ht]
		\centering
		\includegraphics[width=18cm,height=18cm]{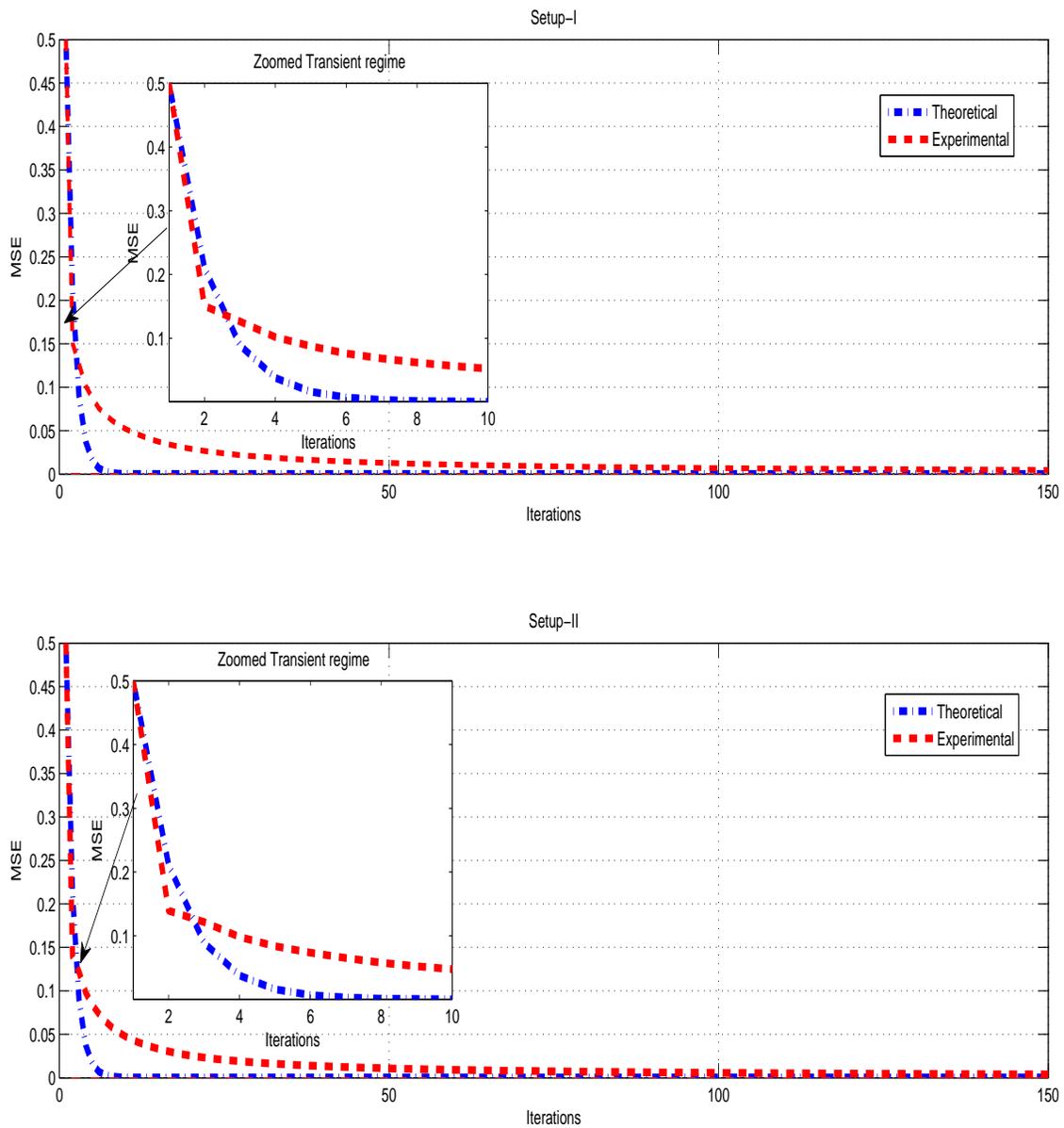}\\
		\caption{Transient behaviour at step-size 0.12, Setup-I: $\textbf{A}=[0.5\hspace{3mm} 0.5;0.5\hspace{3mm} 0.5]$, $\textbf{C} = [0.5\hspace{3mm} 0.5;0.5\hspace{3mm} 0.5]$, Setup-II: $\textbf{A}=[0.666\hspace{3mm} 0.333;0.333\hspace{3mm} 0.666]$, $\textbf{C} = [0.5\hspace{3mm} 0.5;0.5\hspace{3mm} 0.5]$}\label{fig6}
	\end{figure}

	From Fig. \ref{fig6}, we find that the proposed modeling of the transient behaviour of the MSE curve closely matches the experimental transient behaviour for diffusion-KLMS. Please note that the dynamical modeling for the algorithm is more accurate in the transient region of the plots. The transient region is generally specified by the time taken by the MSE plot to decay to $\exp(-1)$ of its initial value \cite{diniz1997adaptive}, which is also called time-constant of the adaptation. We see almost perfect modeling of MSE plot within the range of the time constant.
	
\begin{figure}[!ht]
		\centering
		\includegraphics[width=18cm,height=20cm]{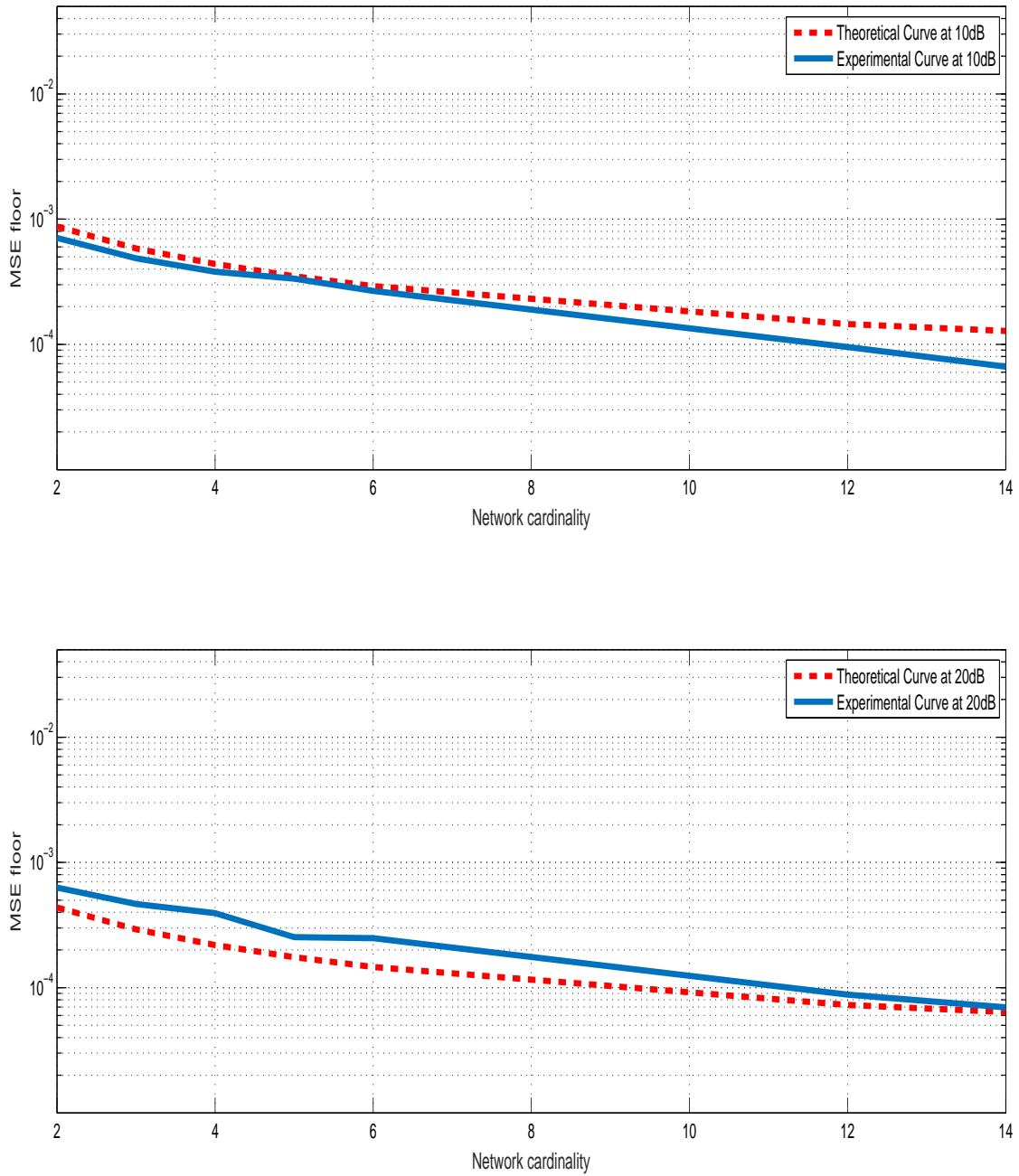}\\
		\caption{Variation of MSE floor with number of nodes for SNR of 10dB and 20dB}\label{fig66}
	\end{figure}

	To study how MSE evolves as we remove or add another node in the network (or in another words change the network size), we plot the experimental MSE floor as a function of network size in Fig. \ref{fig66}. We see that as the network size increases the MSE floor decreases which is an intuitive result. Further, we compare the MSE floor obtained experimentally with the theoretical expression for the same $A,C$ matrices for the given network size. We average over 1000 iterations with various choices of $A$ and $C$, and plot their mean values both for theoretical and experimental MSE floors as a function of the network size.
	We see that the MSE floors as predicted by theoretical expressions derived in Section-V follow the experimentally obtained curves as we increase the size of the network.
	
	\section{Conclusion}
	A new variant of KLMS algorithm has been proposed which is a distributed solution to the non-linear KLMS algorithm. The proposed algorithm converges to a lower MSE floor as compared to the original KLMS algorithm as shown in this paper. Theoretical expressions for both transient and steady-state performance have been derived which closely match with the experimental values. Hence, the proposed diffusion-KLMS is a better adaptive algorithm for estimation as compared to KLMS in distributed non-linear systems. This work has potential applications in non-linear distributed inference over some targeted 5G network's components like detection over massive MIMO and cooperative spectrum sensing for cognitive radio.
	
	
	
\newpage
	\bibliographystyle{IEEEtran}
	\bibliography{paper}

\begin{thebibliography}{10}
\providecommand{\url}[1]{#1}
\csname url@samestyle\endcsname
\providecommand{\newblock}{\relax}
\providecommand{\bibinfo}[2]{#2}
\providecommand{\BIBentrySTDinterwordspacing}{\spaceskip=0pt\relax}
\providecommand{\BIBentryALTinterwordstretchfactor}{4}
\providecommand{\BIBentryALTinterwordspacing}{\spaceskip=\fontdimen2\font plus
\BIBentryALTinterwordstretchfactor\fontdimen3\font minus
  \fontdimen4\font\relax}
\providecommand{\BIBforeignlanguage}[2]{{%
\expandafter\ifx\csname l@#1\endcsname\relax
\typeout{** WARNING: IEEEtran.bst: No hyphenation pattern has been}%
\typeout{** loaded for the language `#1'. Using the pattern for}%
\typeout{** the default language instead.}%
\else
\language=\csname l@#1\endcsname
\fi
#2}}
\providecommand{\BIBdecl}{\relax}
\BIBdecl

\bibitem{boccardi2014five}
F.~Boccardi, R.~W. Heath, A.~Lozano, T.~L. Marzetta, and P.~Popovski, ``Five
  disruptive technology directions for 5{G},'' \emph{Communications Magazine,
  IEEE}, vol.~52, no.~2, pp. 74--80, 2014.

\bibitem{li2012adaptive}
P.~Li and R.~C. De~Lamare, ``Adaptive decision-feedback detection with
  constellation constraints for {MIMO} systems,'' \emph{Vehicular Technology,
  IEEE Transactions on}, vol.~61, no.~2, pp. 853--859, 2012.

\bibitem{wang2014cellular}
C.-X. Wang, F.~Haider, X.~Gao, X.-H. You, Y.~Yang, D.~Yuan, H.~Aggoune,
  H.~Haas, S.~Fletcher, and E.~Hepsaydir, ``Cellular architecture and key
  technologies for 5{G} wireless communication networks,'' \emph{Communications
  Magazine, IEEE}, vol.~52, no.~2, pp. 122--130, 2014.

\bibitem{cattivelli2011distributed}
F.~S. Cattivelli and A.~H. Sayed, ``Distributed detection over adaptive
  networks using diffusion adaptation,'' \emph{Signal Processing, IEEE
  Transactions on}, vol.~59, no.~5, pp. 1917--1932, 2011.

\bibitem{alpaydin2004introduction}
E.~Alpaydin, \emph{Introduction to machine learning}.\hskip 1em plus 0.5em
  minus 0.4em\relax MIT press, 2004.

\bibitem{hayes2009statistical}
M.~H. Hayes, \emph{Statistical digital signal processing and modeling}.\hskip
  1em plus 0.5em minus 0.4em\relax John Wiley \& Sons, 2009.

\bibitem{cattivelli2010diffusion}
F.~S. Cattivelli and A.~H. Sayed, ``Diffusion {LMS} strategies for distributed
  estimation,'' \emph{IEEE Transactions on Signal Processing,}, vol.~58, no.~3,
  pp. 1035--1048, 2010.

\bibitem{lopes2007incremental}
C.~G. Lopes and A.~H. Sayed, ``Incremental adaptive strategies over distributed
  networks,'' \emph{IEEE Transactions on Signal Processing,}, vol.~55, no.~8,
  pp. 4064--4077, 2007.

\bibitem{zhao2012performance}
X.~Zhao and A.~H. Sayed, ``Performance limits for distributed estimation over
  {LMS} adaptive networks,'' \emph{IEEE Transactions on Signal Processing,},
  vol.~60, no.~10, pp. 5107--5124, 2012.

\bibitem{sayed2014adaptive}
A.~H. Sayed, ``Adaptive networks,'' \emph{Proceedings of the IEEE}, vol. 102,
  no.~4, pp. 460--497, 2014.

\bibitem{cattivelli2008diffusion}
F.~S. Cattivelli, C.~G. Lopes, and A.~H. Sayed, ``Diffusion recursive
  least-squares for distributed estimation over adaptive networks,'' \emph{IEEE
  Transactions on Signal Processing,}, vol.~56, no.~5, pp. 1865--1877, 2008.

\bibitem{liu2008kernel}
W.~Liu, P.~P. Pokharel, and J.~C. Principe, ``The kernel least-mean-square
  algorithm,'' \emph{IEEE Transactions on Signal Processing,}, vol.~56, no.~2,
  pp. 543--554, 2008.

\bibitem{haghighat2015variable}
N.~Haghighat, H.~Kalbkhani, M.~G. Shayesteh, and M.~Nouri, ``Variable bit rate
  video traffic prediction based on kernel least mean square method,''
  \emph{IET Image Processing}, 2015.

\bibitem{fan2010proportional}
B.~Fan, W.~Wu, K.~Zheng, and W.~Wang, ``Proportional fair-based joint
  subcarrier and power allocation in relay-enhanced orthogonal frequency
  division multiplexing systems,'' \emph{Communications, IET}, vol.~4, no.~10,
  pp. 1143--1152, 2010.

\bibitem{raja2014adaptive}
M.~A.~Z. Raja and N.~I. Chaudhary, ``Adaptive strategies for parameter
  estimation of box--jenkins systems,'' \emph{IET Signal Processing}, vol.~8,
  no.~9, pp. 968--980, 2014.

\bibitem{slavakis2008sliding}
K.~Slavakis and S.~Theodoridis, ``Sliding window generalized kernel affine
  projection algorithm using projection mappings,'' \emph{EURASIP Journal on
  Advances in Signal Processing}, vol. 2008, no.~1, p. 735351, 2008.

\bibitem{predd2007distributed}
J.~B. Predd, S.~R. Kulkarni, and H.~V. Poor, \emph{Distributed learning in
  wireless sensor networks}.\hskip 1em plus 0.5em minus 0.4em\relax John Wiley
  \& Sons: Chichester, UK, 2007.

\bibitem{honeine2008distributed}
P.~Honeine, C.~Richard, J.~C.~M. Bermudez, and H.~Snoussi, ``Distributed
  prediction of time series data with kernels and adaptive filtering techniques
  in sensor networks,'' in \emph{42nd Asilomar Conference on Signals, Systems
  and Computers, 2008}.\hskip 1em plus 0.5em minus 0.4em\relax IEEE, 2008, pp.
  246--250.

\bibitem{honeine2009functional}
P.~Honeine, C.~Richard, J.~C.~M. Bermudez, H.~Snoussi, M.~Essoloh, and
  F.~Vincent, ``Functional estimation in hilbert space for distributed learning
  in wireless sensor networks,'' in \emph{IEEE International Conference on
  Acoustics, Speech and Signal Processing, 2009. ICASSP 2009.}\hskip 1em plus
  0.5em minus 0.4em\relax IEEE, 2009, pp. 2861--2864.

\bibitem{khalili2012steady}
A.~Khalili, M.~A. Tinati, A.~Rastegarnia, and J.~A. Chambers, ``Steady-state
  analysis of diffusion {LMS} adaptive networks with noisy links,'' \emph{IEEE
  Transactions on Signal Processing}, vol.~60, no.~2, pp. 974--979, 2012.

\bibitem{diniz1997adaptive}
P.~S. Diniz, \emph{Adaptive filtering}.\hskip 1em plus 0.5em minus 0.4em\relax
  Springer, 1997.

\end{thebibliography}
	
	
	
	%
	
	



	\ifCLASSOPTIONcaptionsoff
	\newpage
	\fi

\end{document}